\documentclass[10pt, a4paper]{article}

\usepackage{lrec-coling2024} % this is the new style
\usepackage{booktabs}
\usepackage{svg}
\usepackage{multirow}
\usepackage{float}
\usepackage[T2A,T1]{fontenc}
\DeclareRobustCommand{\cyr}[1]{%
  {\fontencoding{T2A}\selectfont#1}%
}
\usepackage{amsmath}

\title{Setting up the Data Printer with Improved English to Ukrainian Machine Translation}

\name{Yurii Paniv, Dmytro Chaplynskyi, Nikita Trynus, Volodymyr Kyrylov} 

\address{Ukrainian Catholic University, lang-uk initiative, \\
Igor Sikorsky Kyiv Polytechnic Institute,
Università della Svizzera italiana \\
         paniv@ucu.edu.ua, chaplinsky.dmitry@gmail.com, trynus.nikita@lll.kpi.ua, vol@wilab.org.ua \\}

\abstract{
To build large language models for Ukrainian we need to expand our corpora with large amounts of new algorithmic tasks expressed in natural language. Examples of task performance expressed in English are abundant, so with a high-quality translation system our community will be enabled to curate datasets faster. To aid this goal, we introduce a recipe to build a translation system using supervised finetuning of a large pretrained language model with a noisy parallel dataset of 3M pairs of Ukrainian and English sentences followed by a second phase of training using 17K examples selected by k-fold perplexity filtering on another dataset of higher quality. Our decoder-only model named Dragoman beats performance of previous state of the art encoder-decoder models on the FLORES devtest set.
 \\ \newline \Keywords{machine translation, parameter-efficient fine tuning, large language models, unsupervised data selection, perplexity filtering} 
}

\begin{document}

\maketitleabstract

\section{Introduction}

The availability of the data is the most important ingredient when one needs to pretrain general-purpose large language models for a specific natural language task or a set of tasks. While it is relatively easy to obtain a good and balanced dataset under specific domain for the English language, it is much harder to do the same for other under-resourced languages such as Ukrainian.

Since curating a corpus of tasks in Ukrainian is a large endeavor, and given a large body of work done for English, we consider existing instruction tuning datasets as a source of tasks to reuse in Ukrainian using automatic machine translation.

This work focuses on improving the current state of machine translation from English to Ukrainian.

We contribute a recipe for finetuning a large pretrained language model with publicly available data to build a translation system (\autoref{sec:supervised}, \autoref{sec:cleaning}). This  matches state of the art performance of the best encoder-decoder model on a common multilingual benchmark using a consumer GPU with 24 GiB of VRAM.  We release training, evaluation code, datasets, and model at \url{https://github.com/lang-uk/dragoman}. Our main results are summarized in \autoref{tab:final-results}. We provide examples of the top-5 best and worst translations on the FLORES devtest set in the \autoref{sec:examples}.

We base pretrained model selection on evaluation in few-shot learning setting (\autoref{sec:fewshot}). We find that it's a promising method to design tasks without training, and the model can perform comparably to specialized systems given increased inference budget and auxiliary translation scoring functions, yet still underperforms our finetuned recipe.

\begin{table}[]
\centering
\begin{tabular}{ll}
\hline
\textbf{Model} & \textbf{BLEU} $\uparrow$ 
\\
\hline
\textbf{Finetuned} \\
Dragoman P, 10 beams (\autoref{sec:supervised}) & 30.4 \\
Dragoman PT, 10 beams (\autoref{sec:cleaning}) & \textbf{32.3} \\
\hline
\textbf{Zero shot and few shot} (\autoref{sec:fewshot}) \\
Llama 2 7B 2-shot, 10 beams & 20.1 \\
Mistral-7B-v0.1 2-shot, 10 beams & 24.9  \\
gpt-4 10-shot  & 29.5 \\
gpt-4-turbo-preview 0-shot & 30.4 \\
\hline
\textbf{Pretrained encoder-decoder} \\
NLLB-3B, 10 beams & 30.6 \\
OPUS-MT, 10 beams & \textbf{32.2} \\
\hline
\end{tabular}
\caption{Main results. Our Dragoman models improve existing state of the art on translation from English to Ukrainian on FLORES-101 devtest \citep{flores}, a multilingual benchmark of translated sentences from web articles. We compare to state of the art encoder-decoder models,  NLLB-3B \cite{nllbteam2022language} and OPUS-MT \cite{tiedemann-thottingal-2020-opus}. 
}
\label{tab:final-results}
\end{table}

\section{Supervised Finetuning} \label{sec:supervised}

\begin{table*}[t!]
\centering

\resizebox{\textwidth}{!}{%
\begin{tabular}{lrlrlrlr}
\hline
\multicolumn{1}{c}{\multirow{2}{*}{\textbf{Dataset}}} & \multicolumn{1}{c}{\multirow{2}{*}{\textbf{Pairs}}} & \multicolumn{4}{c}{\textbf{Filters}}                                                                           & \multicolumn{1}{c}{\multirow{2}{*}{\textbf{Example Order}}} & {\multirow{2}{*}{\textbf{Best BLEU $\uparrow$}}} \\ 
\multicolumn{1}{c}{}                                  & \multicolumn{1}{c}{}                                    & \textbf{Lang} & \multicolumn{1}{l}{\textbf{BPC}} & \textbf{LaBSE}    & \multicolumn{1}{l}{\textbf{Len diff}} & \multicolumn{1}{c}{}                                   \\ \cline{1-8} 
1m unfiltered                                         & 963k                                                    & -            & -                                   & -                & -                                    & Random   & 28.26          \\
1m filtered                                           & 958k                                                    & En/Uk         & \textless 3.33                       & \textgreater 0.91 & \textless 50                          & Random & 29.47            \\
3m filtered                                           & 2.9m                                                    & En/Uk         & \textless 3.25                       & \textgreater 0.85 & \textless 50                          & By LaBSE score, dissimilar first          & \textbf{30.37}                    \\
8m filtered                                           & 8m                                                      & En/Uk         & \textless 5                          & \textgreater 0.5  & \textless 50                          & By LaBSE score, dissimilar first                  & 30.19             \\ \hline
\end{tabular}%

}

\caption{\label{tab:datasets_paracrawl} Summary of experiments with Paracrawl subcorpora. Legend of filters: \textbf{Lang} denotes language filters, \textbf{BPC} denotes maximum sum of bits per character measures, \textbf{LaBSE} denotes maximum sentence embedding cosine similarity between source and target sentences, \textbf{Len diff} denotes maximum difference in length between source and target in characters. Example ordering impacts data loading in the training loop.}
\end{table*}
We cast machine translation as a likelihood maximization of a density $\operatorname{p}$ of Ukrainian sentences $Y = "\text{\cyr{перекладене речення}}" \in \mathcal{Y}$ conditioned on their English sources with quasi-instruction formatting: $X  = "\text{[INST] translated sentence [/INST]}" \in \mathcal{X}$.

The density is parametrized using a neural network with frozen pretrained weights $\theta$:
\begin{equation}
\operatorname{argmax}_\phi \operatorname{p}_{\theta,\phi}(Y|X)  
\end{equation}

We implement the conditional language modeling objective by masking out tokens of $X$ when computing token-wise cross entropy of shifted targets.
We only optimize extra low rank  adapter \citep{hu2021lora}  parameters $\phi$ after nf4 quantization \citep{qlora}. In practice we use large rank values and adapter mixture weights. All training runs proceed for one epoch and we use dropout \citep{dropout} for regularization against data noise.

We use Mistral-7B-v0.1 \citep{jiang2023mistral} as a base pretrained decoder-only transformer, as it performs favorably in our few-shot experiments (\autoref{sec:fewshot}).

\section{First Phase: Heuristic Filtering of Paracrawl} \label{sec:supervised}

We use the publicly available Paracrawl dataset \cite{banon-etal-2020-paracrawl}. This dataset contains 13,354,365 English-Ukrainian sentence pairs, collected by automatically matching similar sentences in large corpora of internet text.

We have identified issues with translation pairs, including a significant number of repetitive or incorrect examples. We encounter a large subset of repetitive weather forecasts following the template ``The temperature in <x> is <y> degrees,'' and sentences from site navigation menus. Additionally, many texts appear to be scraped from adult websites, containing low-quality, machine-translated samples. We have spotted numerous instances of incomplete or significantly incorrect translation pairs. Some target sentences were written in languages other than Ukrainian.

To control the quality of the sentences, we apply multiple heuristics. 

\vspace{-7pt}
\paragraph*{Language filtering} gcld3 library\footnote{\url{https://github.com/google/cld3}} provides language detection capabilities. We remove all sentences that failed to verify as Ukrainian or English.

\vspace{-10pt}
\paragraph*{Perplexity thresholding} We score source and target sentences using two decoder-only models trained on different monolingual datasets \cite{radford2019language,minixhofer-etal-2022-wechsel} and sum their bits per character measures.

\paragraph*{Translation mismatch filtering} LaBSE \cite{feng2022languageagnostic} embeds sentences into a space, where similar sentences in different languages are close together. We use it to filter out badly aligned sentence pairs.

\paragraph*{Length filtering} The lengths of the original and translated sentences reveal examples that are too short or too long. Absolute differences of lengths point to pairs with long target for the short source and vice versa.

We arbitrarily choose joint values of filtering thresholds to get the desired approximate example counts: 1 million, 3 million and 8 million. We perform multiple experiments with these splits while searching for optimal hyperparameters. We list threshold values in \autoref{tab:datasets_paracrawl} and best results for each subset.

\section{Second Phase: Unsupervised Data Selection on Extended Multi30K}\label{sec:cleaning}

We use the best checkpoint from the previous finetuning phase to train on a high-quality dataset: Extended Multi30K from \citet{saichyshyna-etal-2023-extension}. Switching datasets gives us a performance boost of 1.97 BLEU. We additionally delete 11600 sentences from the dataset using unsupervised perplexity filtering pipeline gaining 0.35 on the dev set that translates to 0.3 BLEU on the devtest subset of FLORES.

\begin{figure}[ht]
    \centering
    \includegraphics[width=0.49\textwidth]{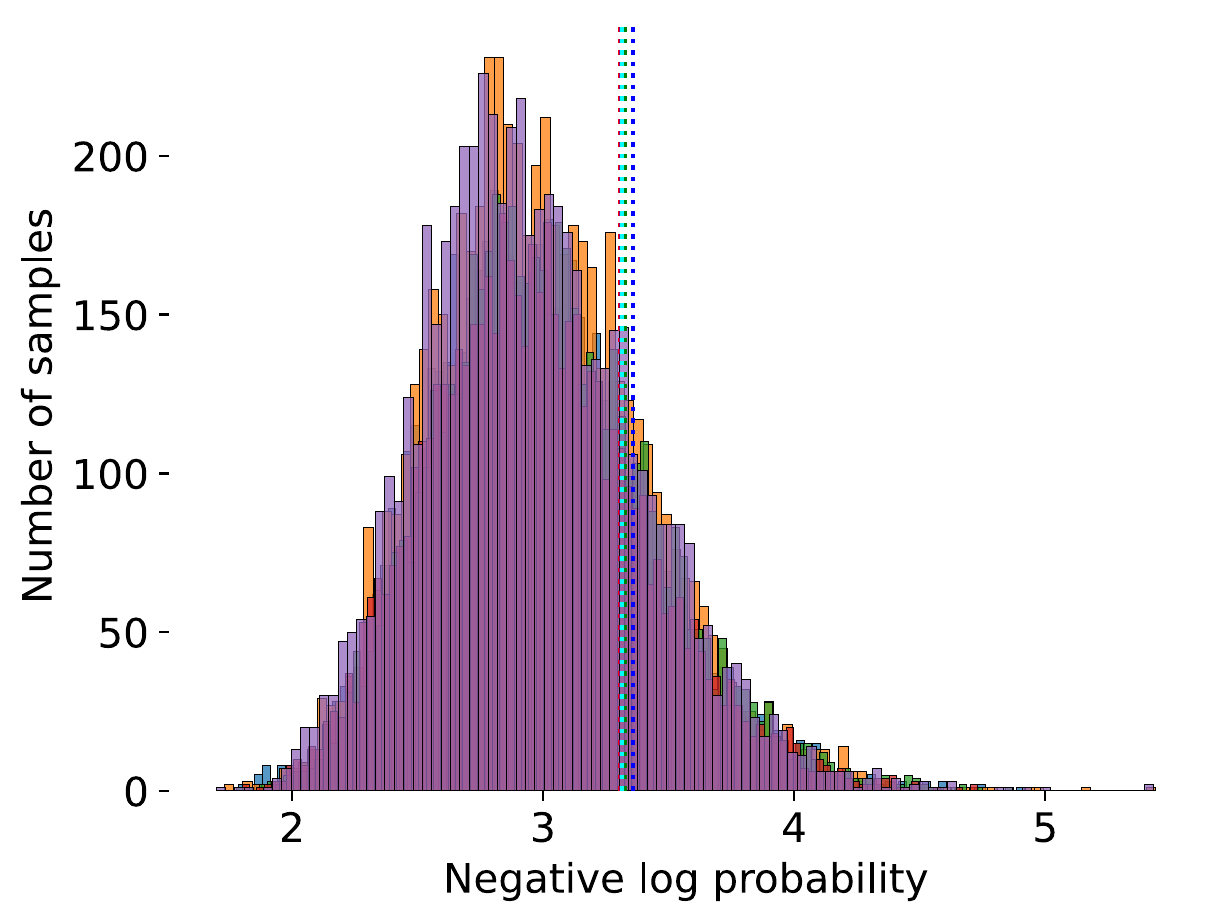}
    \caption{Distributions of sentence log probabilities for each fold superimposed on top of each other. Every bar color represents a unique fold; every vertical line denotes a 60\textsuperscript{th} percentile cutoff threshold. The best percentile is chosen using grid search shown in \autoref{tab:pplthresholds}.}
    \label{fig:log-prob-dist}
\end{figure}

We use perplexity as a data selection criterion to calculate thresholds to filter out highly surprising sentences.
We apply the $k$-fold cross-validation technique to make the perplexity evaluation in-domain. We split the training data into $k=5$ folds and train $k$ models withholding one of the folds from each run. Then we score every sentence using the model that has not seen that sentence in training. Next, we sweep for acceptable threshold values by minimizing BLEU on the development set and report results in \autoref{tab:pplthresholds}. We plot the distribution of scores in \autoref{fig:log-prob-dist}. We also provide threshold sweep results for training from base Mistral-7B-v0.1 checkpoint in \autoref{tab:pplthresholds_scratch}. By comparing finetuned results, we demonstrate that data from the second phase alone is not enough to match the performance of our best checkpoint.

\begin{table}[H]
\centering
\begin{tabular}{lllll}
\hline
\textbf{Threshold} & \textbf{Examples} & \textbf{BLEU} $\uparrow$ &  \\
percentile & & dev & devtest \\
\hline
20\textsuperscript{th} & 5800 & 31.57 & 32.06  \\
40\textsuperscript{th} & 11600 & 31.65 & 32.16  \\
50\textsuperscript{th} & 14500 & 31.76 & 32.36 \\
60\textsuperscript{th} & 17400 & \underline{31.80} & \textbf{32.34} \\
70\textsuperscript{th} & 20300 & 31.51 & 32.17 \\
80\textsuperscript{th} & 23200 & 31.44 & 32.46 \\
95.4\textsuperscript{th} ($2\sigma$) & 28025 & 31.74 & 32.18 \\
Full dataset & 29000 & 31.45 & 32.04 \\

\end{tabular}
\caption{Extended Multi30K log probability thresholds swept on FLORES dev set. We choose the best checkpoint based on model performance on FLORES dev subset using grid search for optimal perplexity threshold value. 
}
\label{tab:pplthresholds}
\end{table}

\section{Few-Shot Translation}\label{sec:fewshot}

Conditioning the model on a sequence of demonstrations of performing some task allows the model to learn this task in-context, also known as ``few-shot learning'' (e.g. \citet{brown2020language}), thanks to the ability of the Transformers to modulate representations of its future tokens using past context, implementing a specialized internal context-dependent learning algorithm inside its weights \citep{vonoswald2023transformers}.

While few-shot learning allows to quickly try any task with a low number of demonstrations,  \citet{liu2022fewshot} have shown that parameter-efficient finetuning allows smaller models achieve better performance, effectively spending less floating point operations per test example at inference time. 

Setting up the model for finetuning requires a lot of work, and in-context learning allows to quickly probe capability of a large model using inference software that performs efficient management of key-value cache for speed \cite{kwon2023efficient}.

To test backbone models before finetuning, we attempt decoding translations with a basic prompt shown in \autoref{fig:prompt}.

\begin{figure}[h]
\cyr{[INST] They are planning to host a party next weekend. [/INST] Вони планують провести вечірку наступного вікенду. \newline
[INST] I enjoy swimming in the ocean and feeling the salty breeze. [/INST] Мені подобається плавати в океані та відчувати солоний вітер. \newline
[INST]}
\caption{Basic 2-shot prompt used for few-shot translation. \texttt{[INST]} prefixes the beginning of the source sentence and \texttt{[/INST]} denotes the beginning of the target translation. These separators are chosen arbitrarily (as in finetuning) and are not special vocabulary items, even though they bear visual resemblance to them.}
\label{fig:prompt}
\end{figure}

We find that the model significantly underperforms compared to current state of the art translation models when using beam search \citep{beamsearch}.

This decoding algorithm performs pruned breadth-first expansion, scoring target sentence prefixes using model's own log probability, approximating maximum a-posteriori estimation of the best translation.

Inspection of the n-best list of translation candidates (beams) reveals that the models can produce high-quality translations, however assign low probabilities to them. We find the best possible translation by rescoring beams using the BLEU score as a loss function \citep{kumar-byrne-2004-minimum} with respect to the reference translation (the so-called ``oracle'').

We employ this oracle rescoring strategy to gauge the potential capability of the model to produce good translations without finetuning, and find that in a regime of increased computation (large width of the beam) and assuming perfect selection capability, a base model is competitive with specialized alternatives. We sweep over a grid of multiple beam widths and report highest attainable BLEU scores in \autoref{tab:oracle}.

\begin{table}[H]
    \centering
    \begin{tabular}{ccc}
        \hline
        \textbf{Beams} & \textbf{Oracle BLEU $\uparrow$} & \\
        & Mistral-7B-v.01 & Llama 2 7B \\
        \hline
        3  & 27.11 & 24.55 \\
        5  & 29.20 & 26.64 \\
        10 & 31.53 & 28.76 \\
        %10 (dyn-dev) & 31.56 \\
        15 & 32.81 & 29.09 \\
        20 & 33.54 & 27.64 \\
        25 & 34.27 & 26.35 \\
        30 & 33.99 & {\small(decoder failure)}\\
        35 & 34.94 & \\
        40 & 34.61 & \\
        \hline
    \end{tabular}
    \caption{We establish the upper bound of the latent capability of pretrained base models to produce high quality translations with by varying beam width on the task of translating sentences from FLORES dev given a 2-shot prompt. The ground truth oracle determines the best beam. We use beam search implementation by \citet{kwon2023efficient} with presence penalty of 0.1.
    The results do not improve monotonically with increasing beam size, and lengths of hypotheses grow with maximum beam size, yielding diminishing returns. This problem can be attributed to label bias \cite{murray-chiang-2018-correcting}, and rectifying it will require extra regularization.}
    \label{tab:oracle}
\end{table}

Consecutive sentences in FLORES are samples from the same document. We hypothesize, dynamically adjusting the prompt by inserting previous translations will improve results. We observe that the model indeed improves translation of certain words such as proper nouns through access to correct definitions provided in the context (\autoref{fig:names}), however its overall performance degrades in other examples.

\begin{figure}

Source:
{\small
 \textsc{\underline{RSPCA} New South Wales chief inspector David O'Shannessy told the ABC that surveillance and inspections of abattoirs should be commonplace in Australia.}}

Hypothesis given random 2-shot context: \cyr{Головний інспектор \underline{\color{red}РСПКА} Нового Південного Уельсу Девід О'Шеннесі повідомив ABC, що спостереження та інспекції аббатств повинні бути звичайним явищем в Австралії.}

Context example: {\small \textsc{[INST] Animal Liberation and the \underline{Royal Society for the Prevention of Cruelty} \underline{to Animals} \underline{(RSPCA)} are again calling for the mandatory installation of CCTV cameras in all Australian abattoirs. [/INST]} \cyr{Організація Звільнення тварин і \underline{Королівське товариство із запобігання} \underline{жорстокому поводженню з тваринами} \underline{(КТЗЖПТ)} знову закликають до обов'язкової установки камер спостереження на всіх австралійських бійнях.}}

Hypothesis given relevant 2-shot context: {\small \cyr{\underline{Головний інспектор Королівського товариства} \underline{із запобігання жорстокому поводженню з тваринами} \underline{\color{blue}(КТЗЖПТ)} Нового Південного Уельсу Девід О'Шеннесі заявив, що спостереження та інспекції бійні повинні бути поширеними в Австралії.}}
\caption{Few-shot translation with contextual prompting allows the model to learn named entities on the fly. Without context, the model makes a wrong guess trying to transliterate the abbreviation.}
\label{fig:names}
\end{figure}

We additionally attempt basic 0-shot with a system prompt \texttt{You translate English sentences into native Ukrainian.}, and 10-shot prompting using automatic prompt selection based on similarity between source sentences experiments with GPT-4 and GPT-4 Turbo and find that commerical systems perform similarly to other open source systems, as shown in \autoref{tab:final-results}.

\section{Discussion and Limitations}\label{sec:discussion}

\paragraph*{Single-sentence translation} Our system is trained on demonstrations of standalone sentence pairs.
\vspace{-7pt}
\paragraph*{Decoder-only models with long context windows} We choose to finetune existing decoder-only models since the choice of models with almost the same architecture but different massive pretraining data is abundant. The number of open-source models released recently and their constant improvement offers a good prospective for the machine translation tasks.

These models receive gradient from all outputs during pretraining, and the self-attention mechanism can see the input, the partial output, and access past examples of translations in its context window using induction heads \citep{olsson2022context}.

For efficiency, we only train on examples with single short sentence pairs and do not pack context windows full of tokens as done in pretraining. In our early experiments, we find that our models still generalize to inputs longer that what is seen in training. This generalization behavior is often attributed to relative position embeddings \cite{xl, csordas-etal-2021-devil}. We leave evaluation of long context attention stability under these conditions for future work.

\paragraph*{Training on the noisy dataset}
Data cleaning has a positive effect on the resulting metrics. However, our models trained on 8 million filtered, examples perform worse than models trained on 3 million examples (\autoref{tab:datasets_paracrawl}).

\paragraph*{Tokenizer performance} 

\begin{figure}[H]
\includegraphics[width=0.49\textwidth]{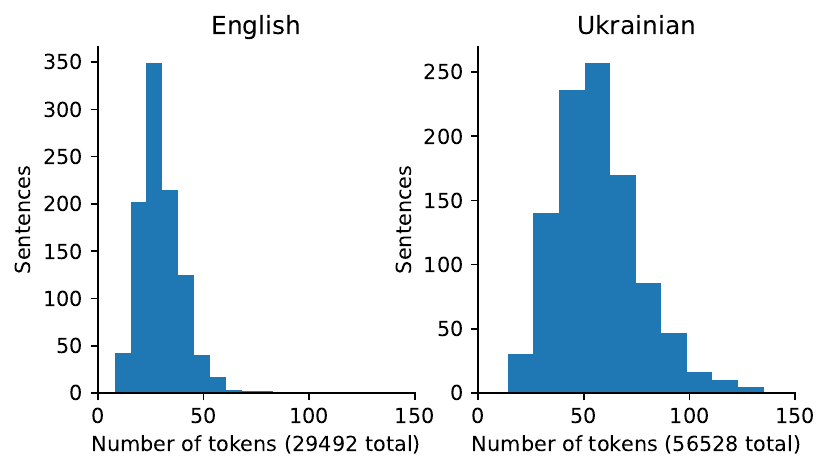}
\caption{Comparison of tokenizer compression rates between English and Ukrainian using the Mistral-7B tokenizer on the FLORES dev set.}
\label{fig:tokens1}
\end{figure}

We used the LLaMA and Mistral tokenizers during our experiments, which use at least twice as many tokens to compress a sentence in Ukrainian of the same length as an English sentence in character. In practice, that means that generating a sentence in Ukrainian takes at least twice as many steps to generate. We show a distribution of sentence token lengths in \autoref{fig:tokens1}.

\paragraph*{Evaluation} We choose BLEU-4 score \cite{bleu} as our core evaluation metric and model selection criterion. BLEU-4 measures 4-gram precisions, where grams are defined as words. We use the implementation and rely on tokenization decisions of \citet{post-2018-call}. This metric is sensitive to minor  differences that do not affect the meaning of the sentence, for example case inflections that tend to cascade to multiple adjacent words. BLEU is known to poorly correlate with human judgement of translation quality, and \citet{freitag-etal-2022-results} recommend learned metrics.

Choosing an appropriate learned metric for judgements of translation quality of Ukrainian requires careful consideration, and incorporation of data informed by the language community, such as a curated corpus of grammar corrections that reflects proper modern use of language \cite{syvokon-etal-2023-ua}.

Regardless of limitations of BLEU, improvement in BLEU still signals improvement in translation quality in our regime.

\paragraph*{WMT22} Our reviewers have pointed out that WMT22 benchmark \cite{wmt22} includes a test set for Ukrainian. Our model achieves 24.72 on the WMT22 test set without any postprocessing, ranking behind the best result of \citet{roussis-papavassiliou-2022-arc} at 25.2 BLEU. We note that the submission that scores relatively low on the WMT22 test, scores comparably to our results on FLORES. These data distribution properties require closer exploration.

\pagebreak
\section{Related Work}

\paragraph*{Translation to Ukrainian} \citet{Maksymenko2023ControllabilityFE,maksymenko-etal-2023-controllability} explore translation controllability by conditioning the model on text embeddings that encode style by finetuning an encoder-decoder model. They claim high quality translations on a private test set.
\vspace{-7pt}

\paragraph*{Instruction-tuned language models} \citet{aya-model} explore large-scale translation efforts to produce a multilingual instruction-tuned language model Aya. This work translates large datasets like the Flan Collection \citep{flan-collection} using the NLLB-3B model  \citep{nllbteam2022language}.

\vspace{-7pt}
\paragraph*{Translation systems} 
\citet{unsupervised-backtranslation} provide an iterated backtranslation recipe to bootstrap neural machine translation systems using generative models: zero-shot translation ability is used to produce candidates for few-shot demonstrations. Filtered few-shot demonstrations are used to sample new sentences for further finetuning for translation in two directions.
\vspace{-7pt}
\paragraph*{Translation benchmarks} Besides FLORES-101 (\citet{flores}, or FLORES-200 \citep{nllbteam2022language}, both include the same data for Ukrainian) dataset used in this work, \citet{tiedemann-2020-tatoeba} provides an additional dataset for multilingual evaluation.
\vspace{-7pt}

\paragraph*{Data selection techniques}
\citet{yang-li-2023-boosting} propose a perplexity filtering pipeline, in which the data is split into k folds to classify low quality augmentation generations produced by surrogate language models. \citet{how-to-train-data-efficient-llms} provide recipes on curating data for language models by directly asking language models to score examples.

\section{Conclusion}
In this work, we build a translation system using a two-phase data cleaning pipeline. We demonstrate matching performance to state-of-the-art encoder-decoder models for English-Ukrainian translation task. Notably, our system exhibits superior performance compared to the NLLB model, which was instrumental in generating the Aya dataset and contributed significantly to the advancement of multilingual language models. Improved machine translation could bring new capabilities to the next generation of large language models trained for the Ukrainian language. The recent improvements made for decoder-only backbones and the general dynamics of this process encourages us: we firmly believe that recipes we propose in this paper can be used to improve the quality of the translation by simply upgrading the backbone model.
\section{Contributions}

Yurii Paniv worked on unsupervised data selection on extended Multi30K dataset, Dmytro Chaplynskyi performed initial training using heuristic filtering of Paracrawl, Nikita Trynus worked on evaluation, Volodymyr Kyrylov designed few-shot learning experiments and evaluation.

\section{Acknowledgements}\label{sec:acknowledgements}
The authors would like to thank Talents for Ukraine project of Kyiv School of Economics for the grant on compute, used for this paper, Ukrainian Catholic University, Roman Kyslyi, and Oleksii Turuta for the fruitful discussions and inspiration. We would also like to thank paper reviewers for their comments.

\section{References}\label{sec:reference}
\bibliographystyle{lrec-coling2024-natbib}
\bibliography{dragoman}

\begin{table*}[p]
\centering
\begin{tabular}{lllll}
\hline
\textbf{Model} & \textbf{BLEU} $\uparrow$  & \textbf{spBLEU} & \textbf{chrF}  & \textbf{chrF++} \\
\hline
\textbf{Finetuned} \\
Dragoman P, 10 beams (\autoref{sec:supervised}) & 30.38 & 37.93 & 59.49 & 56.41 \\
Dragoman PT, 10 beams (\autoref{sec:cleaning}) & \textbf{32.34} & \textbf{39.93} & \textbf{60.72} & \textbf{57.82} \\
\hline
\textbf{Zero shot and few shot} (\autoref{sec:fewshot}) \\
LLaMa-2-7B 2-shot & 20.1 & 26.78 & 49.22 & 46.29 \\
RWKV-5-World-7B 0-shot & 21.06 & 26.20 & 49.46 & 46.46 \\
gpt-4 10-shot  & 29.48 & 37.94 & 58.37 & 55.38 \\
gpt-4-turbo-preview 0-shot & 30.36 & 36.75 & 59.18 & 56.19 \\
Google Translate 0-shot & 25.85 & 32.49 & 55.88 & 52.48 \\
\hline
\textbf{Pretrained} \\
NLLB 3B, 10 beams & 30.46 & 37.22 & 58.11 & 55.32 \\
OPUS-MT, 10 beams & 32.2 & 39.76 & 60.23 & 57.38 \\
\hline
\end{tabular}
\caption{We evaluate generated translations with the sacrebleu library to calculate BLEU, spBLEU, chrF, and chrF++ metrics on the FLORES DEVTEST set. Metric spBLEU was calculated with default BLEU values and tokenizer flores101. Tokenization and detokenization are done using the models’ default tokenizers. Evaluation is performed on detokenized sentences with corresponding reference sentences.}
\label{tab:final-results-extended}
\end{table*}

\begin{table*}[p]
\centering
\begin{tabular}{lllll}
\hline
\textbf{Threshold} & \textbf{Examples} & \textbf{BLEU} $\uparrow$ &  \\
percentile & & dev & devtest \\
\hline
20\textsuperscript{th} & 5800 & 25.14 & 25.49  \\
40\textsuperscript{th} & 11600 & 25.39 & 25.45  \\
50\textsuperscript{th} & 14500 & 25.79 & 25.93 \\
60\textsuperscript{th} & 17400 & \underline{26.07} & \textbf{26.01} \\
70\textsuperscript{th} & 20300 & 26.00 & 25.72 \\
80\textsuperscript{th} & 23200 & 25.90 & 26.08 \\
95.4\textsuperscript{th} ($2\sigma$) & 28025 & 25.91 & 25.81 \\
Full dataset & 29000 & 25.74 & 25.67 \\
\end{tabular}
\caption{Evaluation scores for model, finetuned from Mistral-7B-v0.1 directly on Extended Multi30K dataset. We performed log probability thresholds sweep on FLORES dev set. We demonstrate that data from the second phase alone is not enough to match the performance of our best checkpoint. Perplexity filtering improves downstream performance over training on full Extended Multi30K dataset.  
}
\label{tab:pplthresholds_scratch}
\end{table*}

\appendix

\section*{Appendix}

\section{Translation Examples}\label{sec:examples}

\paragraph{Sample of top 5 worst examples by BLEU from FLORES devtest set}
\textbf{Source:} \textsc{While most of their food would be familiar to us, Romans did have their share of strange or unusual feast items, including wild boar, peacock, snails, and a type of rodent called a dormouse}\\
\textbf{Hypothesis:} \cyr{Хоча більшість їхніх продуктів харчування було б знайомим для нас, римляни мали свою частку дивних або незвичайних страв, включаючи кабана, павича, равликів та гризунів, званих куницями.}\\
\textbf{Reference:} \cyr{Переважно римляни харчувались тим, що знайоме і нам, але були у них і свої дивні та незвичні бенкетні частування, серед яких дикий кабан, павич, слимаки і гризун під назвою вовчок.}\\
\textbf{BLEU:} 1.72
\vspace{5pt}
\\\textbf{Source:} \textsc{"I was moved every time we did a rehearsal on this, from the bottom of my heart."}\\
\textbf{Hypothesis:} \cyr{"Я був зворушений кожного разу, коли ми робили репетицію з цього приводу, від щирого серця".}\\
\textbf{Reference:} \cyr{Мене це торкало кожен раз під час репетиції до глибини душі.}\\
\textbf{BLEU:} 1.91
\vspace{5pt}
\\\textbf{Source:} \textsc{Feral children may have experienced severe child abuse or trauma before being abandoned or running away.}\\
\textbf{Hypothesis:} \cyr{Дитина-дикун, можливо, пережила жорстоке насильство над дітьми або травму, перш ніж бути кинутою або втекти.}\\
\textbf{Reference:} \cyr{Може таке бути, що дикарські діти були об'єктами жорстокого поводження або отримали серйозні травми перед тим, як їх полишили або вони втекли.}\\
\textbf{BLEU:} 2.31
\vspace{5pt}
\\\textbf{Source:} \textsc{Workers must often get their superiors' approval for any decisions they make, and are expected to obey their superiors' instructions without question.}\\
\textbf{Hypothesis:} \cyr{Працівники часто повинні отримувати схвалення своїх начальників на будь-які рішення, які вони приймають, і повинні беззаперечно виконувати вказівки своїх начальників.}\\
\textbf{Reference:} \cyr{Працівникам часто потрібно отримати схвалення керівника щодо будь-яких рішень, а також від них очікують беззаперечного дотримання настанов керівників.}\\
\textbf{BLEU:} 2.32
\vspace{5pt}
\\\textbf{Source:} \textsc{Typically there will be a tuition fee to enroll in these educational programs.}\\
\textbf{Hypothesis:} \cyr{Як правило, буде плата за навчання, щоб зареєструватися в цих освітніх програмах.}\\
\textbf{Reference:} \cyr{Зазвичай такі освітні програми платні.}\\
\textbf{BLEU:} 2.62

\paragraph{Sample of top 5 best examples by BLEU from FLORES devtest set}
\textbf{Source:} \textsc{The East African Islands are in the Indian Ocean off the eastern coast of Africa.}\\
\textbf{Hypothesis:} \cyr{Східноафриканські острови знаходяться в Індійському океані біля східного узбережжя Африки.}\\
\textbf{Reference:} \cyr{Східноафриканські острови знаходяться в Індійському океані біля східного узбережжя Африки.}\\
\textbf{BLEU:} 100.00
\vspace{5pt}
\\\textbf{Source:} \textsc{Earlier the Chinese news agency Xinhua reported a plane to be hijacked.}\\
\textbf{Hypothesis:} \cyr{Раніше китайське інформаційне агентство Сіньхуа повідомило про викрадення літака.}\\
\textbf{Reference:} \cyr{Раніше китайське інформаційне агентство Сіньхуа повідомило про викрадення літака.}\\
\textbf{BLEU:} 100.00
\vspace{5pt}
\\\textbf{Source:} \textsc{For instance, they didn't have corn, nor tomatoes, nor potatoes, nor cocoa, and no ancient Roman ever tasted a turkey.}\\
\textbf{Hypothesis:} \cyr{Наприклад, у них не було ні кукурудзи, ні помідорів, ні картоплі, ні какао, і жоден стародавній римлянин ніколи не скуштував індичку.}\\
\textbf{Reference:} \cyr{Наприклад, у них не було ні кукурудзи, ні помідорів, ні картоплі, ні какао, і жоден стародавній римлянин ніколи не куштував індичку.}\\
\textbf{BLEU:} 90.95
\vspace{5pt}
\\\textbf{Source:} \textsc{The luminosity and rotation are used together to determine a star's Rossby number, which is related to plasma flow.}\\
\textbf{Hypothesis:} \cyr{Світність і обертання використовуються разом для визначення числа Россбі зірки, яке пов'язане з потоком плазми.}\\
\textbf{Reference:} \cyr{Світність і обертання використовуються разом для визначення числа Россбі зірки, яке пов'язане з плазмовим потоком.}\\
\textbf{BLEU:} 83.26
\vspace{5pt}
\\\textbf{Source:} \textsc{But being placed in the "high tropics" just a few degrees north of equator you will need to deal with both heat (always) and strong sun (when the sky is clear, more rarely).}\\
\textbf{Hypothesis:} \cyr{Але перебуваючи в "високих тропіках" всього в декількох градусах на північ від екватора, вам доведеться мати справу як з спекою (завжди), так і з сильним сонцем (коли небо чисте, рідше).}\\
\textbf{Reference:} \cyr{Але, перебуваючи в "високих тропіках" всього в декількох градусах на північ від екватора, вам доведеться мати справу як зі спекою (завжди), так і з палючим сонцем (коли небо чисте, рідше).}\\
\textbf{BLEU:} 82.47

\end{document}